\documentclass[10pt]{article}
\usepackage[legalpaper, margin=1in]{geometry}
\usepackage{amsmath,amsfonts}
\usepackage{algorithmic}
\usepackage{array}
\usepackage[caption=false,font=normalsize,labelfont=sf,textfont=sf]{subfig}
\usepackage{textcomp}
\usepackage{stfloats}
\usepackage{url}
\usepackage[ruled,vlined]{algorithm2e}
\usepackage{adjustbox}
\usepackage{verbatim}
\usepackage{wrapfig,lipsum,booktabs}
\usepackage{graphicx}
\usepackage{cite}
\usepackage{xcolor}
\usepackage[subfigure]{tocloft}
\begin{document}


\title{CEU-Net: Ensemble Semantic Segmentation of Hyperspectral Images Using Clustering}

\author{Nicholas Soucy and Salimeh Yasaei Sekeh}%

\date{}

\maketitle

\begin{abstract}
Most semantic segmentation approaches of  Hyperspectral images (HSIs) use and require preprocessing steps in the form of patching to accurately classify diversified land cover in remotely sensed images. These approaches use patching to incorporate the rich neighborhood information in images and exploit the simplicity and segmentability of the most common HSI datasets. In contrast, most landmasses in the world consist of overlapping and diffused classes, making neighborhood information weaker than what is seen in common HSI datasets. To combat this issue and generalize the segmentation models to more complex and diverse HSI datasets, in this work, we propose our novel flagship model: Clustering Ensemble U-Net (CEU-Net). CEU-Net uses the ensemble method to combine spectral information extracted from convolutional neural network (CNN) training on a cluster of landscape pixels. Our CEU-Net model outperforms existing state-of-the-art HSI semantic segmentation methods 
and gets competitive performance with and without patching when compared to baseline models. We highlight CEU-Net's high performance across Botswana, KSC, and Salinas datasets compared to HybridSN~\cite{c12} and AeroRIT~\cite{c22} methods. 

\end{abstract}

\section{Introduction}
Between climate change, invasive species, and logging enterprise, it is important to know which ground-types are where on a large scale. Recently, due to the wide spread use of satellite imagery, hyperspectral images (HSI) are available to be utilized on a grand scale in ground-types semantic segmentation\cite{cook2013nasa, garcia2022efficient, yuan2021review,c10}. Classifying ground-types gives information that can reduce the ecological impact of logging enterprise to allowing researchers to prevent Ash trees from getting affected by the Emerald Ash Borer\cite{c1}. 

Ground-type semantic segmentation is a challenging problem in HSI analysis and the remote sensing domain. Ground-types in a natural forest environment are overlapping, diverse, and diffused. In contrast, the two most common datasets, Indian pines and Salinas \cite{c8} datasets are small, and land-separated. Due to the already segmented nature of farmland and small sample size, the techniques that apply to these datasets do not translate well to large complex natural forest. In contrast, recent advancements in remote sensing imaging has increased spectral resolution exponentially which affects the segmentation models' performance significantly\cite{c12}. Therefore, models that exploit the rich spectral information more efficiently can see higher accuracy without a large performance cost.

\subsection{Patching}

\begin{figure*}[t]
\vspace{4mm}
\centering
\includegraphics[scale=0.47]{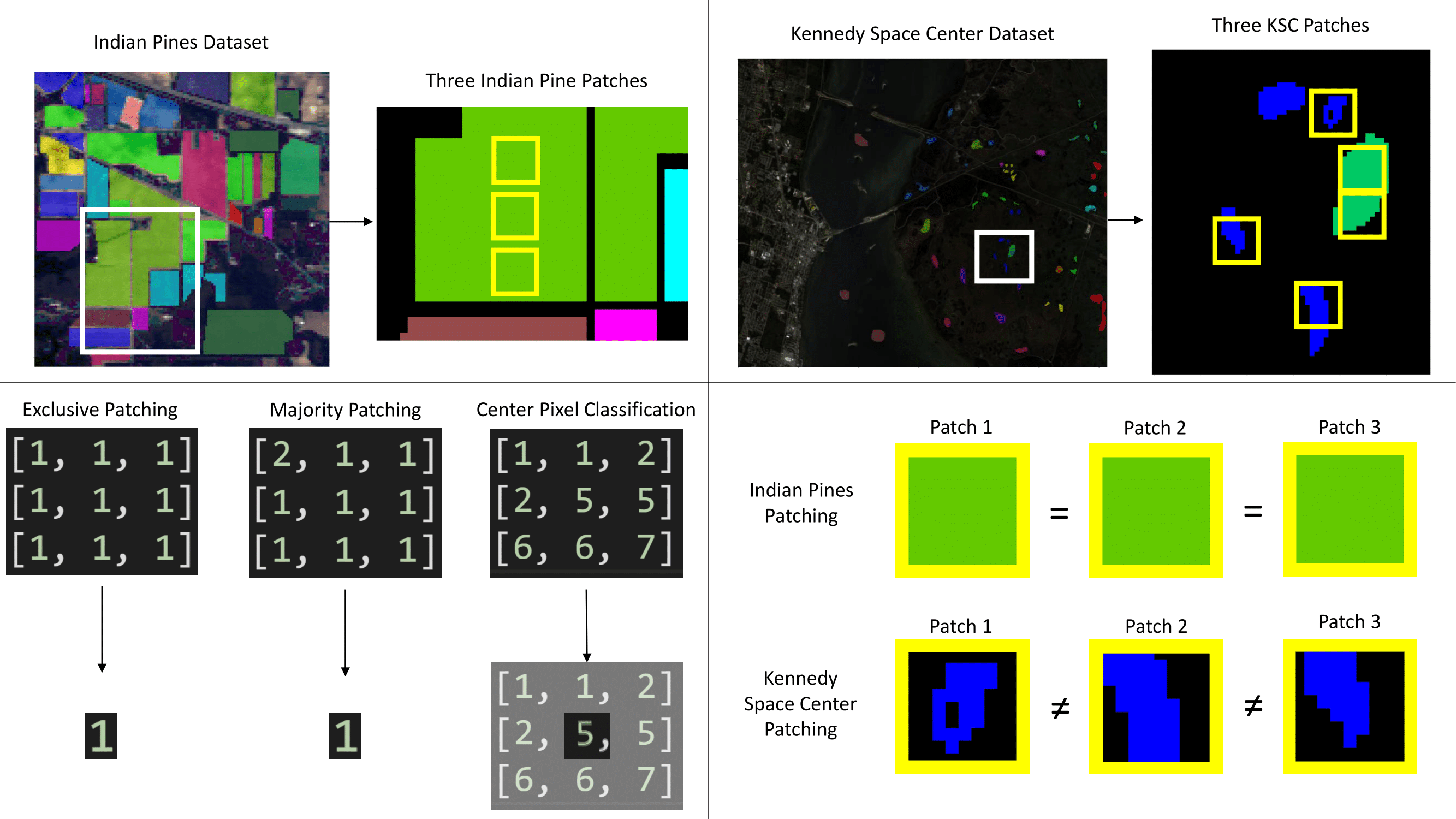}
\caption{
Top left shows a zoomed in area of the Indian Pines dataset with three example patches created during the patching process with the same center pixel class. Top right shows a zoomed in area of the Kennedy Space Center dataset with five example patches.
Bottom right shows the three types of patching  1) Exclusive patching takes a patch of $n$ x $n$ pixels and reduces the size of the dataset by downsizing the patch into one pixel if all classes in the patch match, similar to convolution. 2) Similar to exclusive patching, majority patching will downsize the patch into one pixel based on the the most popular class in that patch. Both exclusive and majority patching are not used in our experiments and other works due to the already small number of labeled pixels. However, we include them as they could be used in future datasets which have potentially millions of labeled pixels. 3) Center pixel creates a $n$ x $n$ patch for each pixel that contains all the neighborhood information of that pixel as input into the CNN. Then it classifies the center pixel in each patch. Farmland datasets like Indian Pines have better neighborhood information than a diffused forest, and therefore benefit more heavily from center pixel classification. Datasets like Kennedy Space Center have smaller neighborhood information and CPC has little impact on overall test accuracy\cite{c24} as shown in Tables \ref{no-patching} and \ref{patching}.
Bottom right shows how the three patches with the same center pixel class from the Indian Pines have identical neighbors, this shows the high value of the neighborhood information, therefore patching would be a useful step to improve semantic segmentation accuracy. However, in contrast the patches in the Kennedy Space Center dataset do not have as similar looking neighbors and therefore the neighborhood information is not as useful. }
\label{patching_over}
\end{figure*}

Patching is an practical preprocessing technique which often increases overall test accuracy of a semantic segmentation model by using spatial neighborhood information via overlapping patches\cite{c2,c12,c14,c15,c22}. Patching is implemented by three approaches: exclusive, majority, and center pixel classification. Examples of these patching techniques are described in Figure \ref{patching_over}. Despite patching improving the performance of segmentation models with particular datasets like Indian Pines and other farmland datasets\cite{c22}, it is often not as useful in datasets that have diverse overlapping classes as shown in Tables \ref{no-patching} and \ref{patching}. Due to the limited number of labeled samples and the nature of individual pixel classification, exclusive and majority patching are rarely used in hyperspectral semantic segmentation models because these techniques would further reduce the dataset size. In addition, exclusive and majority patching would not be possible in datasets with diverse and overlapping classes. Therefore, we focus on center pixel classification in the following sections.

Center pixel classification (CPC) is used in more recent works including one of our baseline models HybridSN~\cite{c12}. The CPC method is implemented by taking a patch of size $n$ x $n$ for each pixel in the dataset as input to the model to capture the spatial neighbors of the pixel. This exponentially increases the time complexity of training and testing due to each sample being an $n$ x $n$ x $b$ patch, where $b$ is the number of spectral bands, instead of just a single pixel with spectral bands $b$. This technique can work for many datasets were other techniques like exclusive and majority patching will not because the size of the dataset is not reduced, and datasets with overlapping classes can still classified. This can lead to a dramatic increase in time complexity with diminishing returns to test accuracy if the neighborhood information is not as useful. Figure \ref{patching_over} bottom left visually demonstrates CPC. 

There has been a push in recent works that focus on neighborhood information instead of spectral information to further increase semantic segmentation accuracy in these popular HSI datasets~\cite{c3, c11, c12, he2017multi, c14, c15}. Due to industrial farming techniques, corn is grown in a single patch, therefore a pixel of corn will be accompanied by other corn pixels. This insures that the neighbors of each pixels in a single class are all similar. This information can be used in the classification network to much success. However, once most land types many HSI researches are interested in have diverse overlapping classes, neighborhood information is weak. Datasets like Kennedy Space Center and Botswana are the closest examples of this phenomenon with a sufficient number of labeled pixels (see Figure \ref{patching_over}). These two datasets are left out of many works due to their small labeled sample size. However, we include these datasets to compare and contrast the performance of existing patching-based methods and highlight the weakness in their assumptions. The goal is to develop semantic segmentation models that are dataset independent and provide competitive performance without implementing patching as preprocessing step. 

\subsection{Semantic Segmentation}
 Among several segmentation models~\cite{c12,c22,c23,c24}, to our knowledge the most successful CNN-based model for popular HSI datasets is HybridSN~\cite{c12}. Instead of using exclusively 3D-CNNs and sacrificing runtime, or using exclusively 2D-CNNs and sacrificing accuracy, they propose a hybrid spectral CNN (HybridSN) for hyperspectral semantic segmentation\cite{c12}.

In general, the HybridSN is a spectral-spatial 3D-CNN followed by spatial 2D-CNN. The 3D-CNN facilitates the joint spatial-spectral feature representation from a stack of spectral bands. The 2D-CNN on top of the 3D-CNN further learns more abstract-level spatial representation via neighborhood information. Moreover, the use of hybrid CNNs reduces number of parameters in the model compared to the use of 3D-CNN alone. 

Semantic segmentation has seen great strides in the medical field with the introduction of a novel deep neural network (DNN) architecture called U-Net~\cite{c17,c18,c21}. The idea to use U-Net for semantic segmentation in HSI has to our knowledge been done only once from the paper AeroRIT~\cite{c22}. The novelty in their U-Net architecture adds complexity via a custom squeeze and excitation block. However, with a high number of features, the time complexity increases exponentially. In addition, AeroRIT did not include studies on other datasets and they used neighborhood information in the form of patching as a preprocessing step. 

To combat these issues in HSI semantic segmentation, we increase the effectiveness of U-Net with our novel Clustering Ensemble U-Net (CEU-Net) by using an ensemble method to create separate parallel models that are trained in subsets of pixels for better overall classification accuracy.

\noindent To summarize, our contributions in this paper are,
\begin{enumerate}
    \item Debuting Clustering Ensemble U-Net (CEU-Net) for HSI semantic segmentation to get more competitive accuracies without neighborhood information.
    \item Empirical analysis on the common preprocessing technique of patching and focusing more on spectral information instead of neighborhood information to make our model, CEU-Net, more data independent.
    \item Creating a strong single U-Net for better semantic segmentation accuracy in common HSI datasets.
\end{enumerate}

\section{Related Works}
Current machine learning (ML) based solutions employing neural networks focus on semantic segmentation. Due to the lack of sufficient labeled samples in popular HSI datasets, this is often treated as a pixel classification problem.

Recent works have been using 2D and 3D CNN in both feature reduction and semantic segmentation techniques in order to implement neighborhood information in addition to spectral information\cite{c2,c12,he2017multi,c14,c23}. The works focus on 2D CNN architectures\cite{c11} are older, however, more recent works have focused on 3D CNN architectures or 2D-3D hybrids with greater success\cite{c12,he2017multi,c14,c15}.

Several works including \cite{c2,c12,c14,c23,c24} employ a combination three datasets: Indian Pines, Salinas, and Pavia University due to their well labeled nature and easy access. We will be focusing on these datasets in addition to Kennedy Space Center and Botswana\cite{c8}.

\subsection{Neighborhood Information}

The use of neighborhood information is not new in HSI semantic segmentation, almost all of the CNN models for HSI semantic segmentation use neighborhood information in the form of patching as a preprocessing step\cite{c3,c11,c12,he2017multi,c14,c15}. Models use neighborhood information due to the nature of the most popular HSI datasets: Indian Pines, Salinas, and Pavia University. These flagship datasets are popular due to their consistent use and number of labeled pixels. However, the vast majority of HSI images are of dense forest areas with diverse ground-types, but are not labeled\cite{c2, c20}.

In \cite{c24}, the authors discuss patching and its shortcomings by demonstrating how patching only exploits the local spatial information and results in high noise in the data when classes overlap frequently. They propose a full patching network called SPNet with a end-to-end deep learning architecture to do the spectral patching instead of manual analysis. However, SPNet is still a network based approach that adds significant training time to semantic segmentation over the common patching method CPC. Further, this work shows how patching is not always the best approach to semantic segmentation. Therefore, we do not include this as in this paper we focus on improving solely spectral information in our semantic segmentation network for datasets that are more complex like tree species data. 

\begin{figure*}[t!]
\vspace{4mm}
\centering
\includegraphics[scale=0.35]{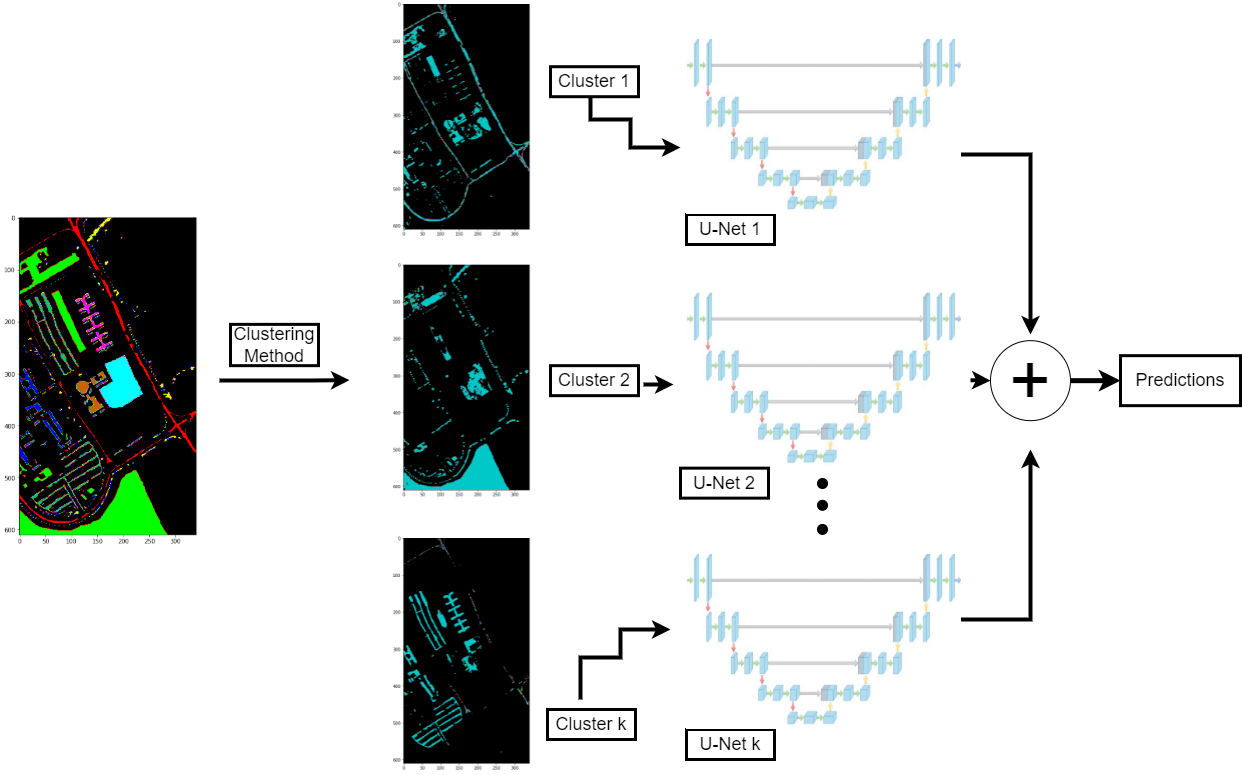}
\caption{This figure is the holistic overview of our clustering ensemble U-Net model (CEU-Net). We first choose a clustering method and $k$ cluster number that is optimal for each dataset based on preliminary experiments. After the clustering method is trained on our training data, we train $k$ U-Nets, one for each cluster. Then we clustering our test data and send each cluster into their respective U-Nets. Then we concatenate the $k$ U-Net predictions as the overall model accuracy.)}
\label{CE U-Net}
\end{figure*}

\subsection{Semantic Segmentation}

{\it U-Net Model}: AeroRIT included a U-Net architecture that added complexity via a custom squeeze and excitation block. This is a common practice in the RGB image domain. It works by scaling network responses by modeling channel-wise attention weights, similar to the residual layer in ResNet\cite{c21}. The authors use this on large-scale hyperspectral data, however, with the number of channels (bands) that are in hyperspectral compared to RGB, the time complexity increases exponentially. In addition, AeroRIT did not include studies on other datasets and the neighborhood information is used in the form of patching as a preprocessing step.

\section{Clustering Ensemble U-Net (CEU-Net)}

One single U-Net is a strong architecture for semantic segmentation, however, without neighborhood information it is difficult to get competitive accuracy versus models that use it. Our solution to this challenge is our proposed Clustering Ensemble U-Net model. In machine learning the ensemble technique is used to improve the accuracy and stability of learning models, especially for the generalization ability on complex datasets. The overview of our Clustering Ensemble U-Net model is demonstrated in Figure \ref{CE U-Net}. We propose a novel technique for separating dissimilar pixels by performing unsupervised clustering on pixels via their spectral signature. We formalize our proposed model as follows:
\subsection{Notation} For an HSI semantic segmentation problem, conditioned on an observed image $\mathbf{x}\in\mathbb{R}^{N\times w}$ with $N$ pixels and $W$ spectral range. The objective is to learn the true posterior distribution $p(\mathbf{y}|\mathbf{x})$, where $\mathbf{y}\in\{1,\ldots m\}^N$, and $1,\ldots,m$ are land type labels. Throughout the paper we use the notations below:
\begin{itemize}
    \item $\{x_i,y_i\}_{i=1}^N$: Training data where $x_i$ is a pixel and $y_i$ is label, $y_i\in\{1,\ldots,m\}$. 
    \item Classifier $F$: A function mapping the input space $\mathcal{X}$ to a set of labels $\mathcal{Y}$, i.e. $F: \mathcal{X}\mapsto \mathcal{Y}$. In this paper this map function is a U-Net model, $F=F^{U-Net}$. 
    \item $\mathcal{L}_{\theta_j}(F_j(\mathbf{x}),y)$: Loss function with parameter set $\theta_j$. In U-Net model, $F_j^{U-Net}$, the parameter set $\theta_j$ is the network's weight matrix and offsets. 
    \item $\mathbf{\omega}$: Ensemble weight vector, $\mathbf{\omega}=[\omega_1,\ldots,\omega_k]^T$. 
\end{itemize}
\subsection{Methodology}
Training a classifier is performed by minimizing a loss function: 
\begin{equation}
    \widehat{\theta}=\arg\min\limits_{\theta} \mathcal{L}_\theta(F(\mathbf{x}),y).
\end{equation}
In ensemble approach with $k$ classifier, $F_1(\mathbf{x}),\ldots F_k(\mathbf{x})$ and weight vector $\omega=[\omega_1,\ldots,\omega_k]^T$ satisfying $\sum\limits_{j=1}^k \omega_k=1$, we find the optimal parameter set $\widehat{\theta}$ as follows:
\begin{equation}\label{eq:ensemble-loss}
  \widehat{\theta}=\arg\min\limits_{\theta, k,\omega} \sum\limits_{j=1}^k \omega_j \mathcal{L}_{\theta_j}(F_j(\mathbf{x}),y).   
\end{equation}
Our proposed CEU-Net architecture extends (\ref{eq:ensemble-loss}) by utilizing clustering method: Let $C_1(\mathbf{x}),\ldots,C_k(\mathbf{x})$ be $k$-clusters of training data $\{x_i,y_i\}_{i=1}^N$ with label sets $y_{C_1},\ldots,y_{C_k}$, respectively. CEU-Net optimizes parameter set $\theta$ by
\begin{equation}\label{eq:ensemble-loss2}
  \widehat{\theta}=\arg\min\limits_{\theta, k,\omega} \sum\limits_{j=1}^k \omega_j \mathcal{L}_{\theta_j}(F_j(C_j(\mathbf{x})),y_{C_j}).      
\end{equation}
Note that in CEU-Net model $F_j$ is a single U-Net model i.e. $F_j=F_j^{U-Net}$. A pseudocode of our CEU-Net model is illustrated in Algorithm~\ref{alg:CEU-Net}.

\begin{algorithm}
\SetAlgoLined
{\bf Input:} HSI Data $\{x_i,y_i\}_{i=1}^N$ \\
{\bf Output:} Overall Test Accuracy\\
Set $k$ to be the number of clusters \\
Set $T$ to be the number of trials\\
Determine $\omega_{1}, \dots, \omega_k$ to be the ensemble weights\\
 \For{$t = $ 1, \dots, T}
 {
Cluster training data $\{x_i,y_i\}_{i=1}^N$ as $C_1(\mathbf{x}),\ldots,C_k(\mathbf{x})$ with label sets $y_{C_1},\ldots,y_{C_k}$\\
 \For{$j = $ 1, \dots, k}
 {
Train $F^{U-Net}_j$ using data points in $j$th cluster  $\{C_j(\mathbf{x}),y_{C_j}\}$ and loss function $\omega_j\mathcal{L}_{\theta_j}$\\
Store test accuracy $AC_{tj}$
}
Sum $AC_{tj}$ i.e ${AC}_t=\sum\limits_{j=1}^k AC_{tj}$
 }
Compute the average of $\{{AC}_1,{AC}_2,\ldots,{AC}_T\}$ i.e. 
\begin{center}
$\frac{1}{T}\sum\limits_{t=1}^T{AC}_t\rightarrow AC$.
\end{center}
\vspace{-0.5cm}
Report ${AC}$\\
\caption{CEU-Net Algorithm}
\label{alg:CEU-Net}
\end{algorithm}

In the practical implementation of Algorithm~\ref{alg:CEU-Net} the value of weights $\omega=[\omega_1,\omega_2,\ldots,\omega_k]^T$ is determined experimentally.
 We then take the training data and train an unsupervised clustering method that separates the pixels into $k$ clusters. Both $k$ and the clustering method will be tuned for each dataset. We then send the training pixels from each cluster into $k$ separate U-Nets for separate training in a supervised fashion with categorical cross entropy as the loss function. This way, each U-Net becomes an expert in their given cluster. After each U-Net is trained, we use the trained clustering method to cluster the testing data into $k$ clusters. Then we predict the labels for each cluster using the corresponding trained U-Net for each cluster. Finally, the U-Nets' predicted labels are concatenated and we compare it to the ground-truths for overall testing accuracy. Each U-Net is identical to the single U-Net architecture.

\section{Experimental Results}

The experimental results section is divided into two main parts, the first discusses the performance of CEU-Net in the context of the state-of-the-art semantic segmentation algorithm and illustrates key insights into the expected behavior of CEU-Net. 
The second part emphasizes the efficiency improvement of CEU-Net and hyper-parameter tuning. 
\subsection{Setup}
We briefly outline the datasets, feature extraction, U-Net architecture, configuration, clustering methods, and metrics used across our experiments. 

\paragraph{Data Sets}

In this experiment, we choose five datasets: Indian Pines, Salinas, Pavia University, Kennedy Space Center, and Botswana~\cite{c8}. HSI data is infamously difficult to label due to the professional and time requirements necessary to label ground-types\cite{c19}. These well known HSI datasets are well labeled and will provide good testing data for our semantic segmentation techniques.

These datasets while used profusely in the ML hyperspectral community, have quite a few flaws.
\begin{enumerate}
    \item {\it Easily Segmentable}: Indian Pines, Salinas, and Pavia University are either farmland, or a campus. These land areas are quite separable spatially. This means grass is often next to other grass and tar is next to other tar etc. This makes training an easy task in just the pixel domain.
    \item {\it Not Representative of Most Land Areas}: A vast majority of land in the world is forest regions and most hyperspectral remote sensing is done in these areas\cite{c20}. Therefore, the existing semantic segmentation models for HSI in remote sensing are not transferable to other landscapes due to the unavailability of labeled samples.
    \item {\it Small Amount of Labeled Samples}: Due to the difficulty of labeling HSI data, the amount of pixels in a dataset is often not a good description of its entire size. All the datasets we use here have a labeled pixel percentage under 50\% as shown in Table \ref{ds-info}. This could possibly lead to over-fitting when presented with complex architectures. 
\end{enumerate}

\begin{table}
\begin{center}
\scalebox{0.7}{%
\begin{tabular}{lllllll}
\hline
Dataset          & Sensor    & Spectral Bands & Number of Classes & Number of Pixels & Number of Labeled Pixels & Percent of Labeled Pixels \\ \hline
Indian Pines     & AVIRIS    & 200            & 16                & 21,025           & 10,249                   & 48.75\%                   \\
Salinas          & AVIRIS    & 204            & 16                & 111,104          & 54,129                    & 48.71\%                   \\
Pavia University & ROSIS     & 103            & 9                 & 207,400          & 42,776                   & 20.62\%                   \\
KSC              & AVIRIS    & 176            & 13                & 314,368          & 5,211                    & 0.017\%                   \\
Botswana         & NASA EO-1 & 145            & 14                & 377,856          & 3,248                    & 0.009\%                   \\ \hline
\end{tabular}}
\vspace{0.2cm}
\caption{\label{ds-info}The table shows the information of most popular datasets in HSI semantic segmentation~\cite{c8}.}
\end{center}
\end{table}

It is clear why the first three datasets are picked more often, they have a larger amount of labeled pixels. All of these datasets have large, separated regions for their ground truths and not more pixel specific classes like tree species, making neighborhood information a smart choice to increase semantic segmentation accuracy for these specific datasets. A new dataset called AeroRIT\cite{c22} is introduced that has more labeled pixels, however, because it (1) does not have diverse classes, (2) has small number of classes, and (3) is not practical for forest remote sensing, we did not include it in our study.

\paragraph{Feature Extraction}

To reduce the dimensionality and for feature extraction, in this paper we use PCA as our baseline feature extraction to compare our other two customized CNN-based techniques. We apply autoencoder models using customized 2D and 3D convolutional auto encoder architectures for dimensionality reduction.

Many related works have shown that 2D and 3D CNN structures have had success when compared to traditional feature extraction techniques\cite{c2, c12}. Therefore, to start off our first autoencoder architecture we decided to use a 2D convolutional autoencoder. This way, if the accuracy produced by the 2D autoencoder are sufficient, we do not have to apply a customized 3D auto encoder which would be more computationally expensive  to train.

To customize both the 2D and 3D convolutional autoencoders, we vary the kernel/pooling size and strides  to determine the efficient  and optimal dimensionality to train U-Nets. However, the operations, input shapes, and activation functions are kept constant. The autoencoders have the exact same layers except each 2D layer is its 3D equivalent in the 3D autoencoder. At the end of the decoder network, we have an upscaled image of the same size as the original to compare to for unsupervised learning. The loss function used is Mean Squared Error.

\paragraph{Single U-Net Architecture}

For our main clustering ensemble model contribution we develop a CNN based model for semantic segmentation that is lightweight to deal with the large number of features per sample. Based on this strategy, we propose a custom CNN that focuses on the rich spectral data available for each pixel, therefore customized CNN under a U-Net backbone was our first choice among various architectures.

A general U-Net consist of two parts:  a contracting path (left side of 'U') and an expansive path (right side of 'U'). U-Net's novelty is in supplementing a usual contracting network by successive layers where the typical pooling layers are replaced by upsampling. This technique increases the resolution for each pixel. The successive convolutional layer can then learn to assemble a precise output based on this information. In addition, U-Net has a large number of feature channels in the upsampling part, which allow the network to propagate context information to higher resolution layers. This makes the expansive path symmetric to the contracting path yielding the famous 'U'-shaped architecture.

For our specific U-Net architecture, the contracting path consist of three 3x3 2D convolutions followed by a leaky rectified linear unit (LReLU) then a Dropout layer with a 20\% rate to prevent over-fitting. For our expansive path we have three 3x3 convolutions 2D transposes with the last layer outputting a logit array of size equal to the number of classes in the dataset. We then do a softmax layer for calculating semantic segmentation accuracy. This architecture is based on the original U-Net and a github\cite{c16,c21}. Table \ref{params} shows an example of the layer-wise summary of our single U-Net for the Pavia University dataset with no patching. When patching is added, the patch size replaces the $1$ x $1$ output shapes in each layer with $n$ x $n$ where $n$ is the patch size. Therefore our single U-Net and CEU-Net can work with various patching techniques as well.

\begin{table}[t]
\centering
\begin{tabular}{lll}
\hline
Layer (Type)                                                                           & Output Shape & \# Parameter \\ \hline
input\_1 (InputLayer)                                                                  & (1,1,30)     & 0            \\
conv1\_2 (Conv2D)                                                                      & (1,1,64)     & 17280        \\
\begin{tabular}[c]{@{}l@{}}batch\_normlization\\ (BatchNormalization)\end{tabular}     & (1,1,64)     & 256          \\
leaky\_re\_lu (LeakyReLU)                                                              & (1,1,64)     & 0            \\
dropout (Dropout)                                                                      & (1,1,64)     & 0            \\
conv2\_2 (Conv2D)                                                                      & (1,1,128)    & 73728        \\
\begin{tabular}[c]{@{}l@{}}batch\_normalization\_1\\ (BatchNormalization)\end{tabular} & (1,1,128)    & 512          \\
leaky\_re\_lu\_1 (LeakyReLU)                                                           & (1,1,128)    & 0            \\
dropout\_1 (Dropout)                                                                   & (1,1,128)    & 0            \\
conv3\_2 (Conv2D)                                                                      & (1,1,256)    & 294912       \\
\begin{tabular}[c]{@{}l@{}}batch\_normlization\_2\\ (BatchNormalization)\end{tabular}  & (1,1,256)    & 1024         \\
leaky\_re\_lu\_2 (LeakyReLU)                                                           & (1,1,256)    & 0            \\
dropout\_2 (Dropout)                                                                   & (1,1,256)    & 0            \\
deconv3 (Conv2DTranspose)                                                              & (1,1,256)    & 589824       \\
\begin{tabular}[c]{@{}l@{}}batch\_normalization\_3\\ (BatchNormalization)\end{tabular} & (1,1,256)    & 1024         \\
leaky\_re\_lu\_3 (LeakyReLU)                                                           & (1,1,256)    & 0            \\
dropout\_3 (Dropout)                                                                   & (1,1,256)    & 0            \\
concatenate (Concatenate)                                                              & (1,1,384)    & 0            \\
deconv2 (Conv2DTranspose)                                                              & (1,1,128)    & 442368       \\
\begin{tabular}[c]{@{}l@{}}batch\_normalization\_4\\ (BatchNormalization)\end{tabular} & (1,1,128)    & 512          \\
leaky\_re\_lu\_4 (LeakyReLU)                                                           & (1,1,128)    & 0            \\
dropout\_4 (Dropout)                                                                   & (1,1,128)    & 0            \\
concatenate\_1 (Concatenate)                                                           & (1,1,192)    & 0            \\
deconv1 (Conv2DTranspose)                                                              & (1,1,9)      & 15561        \\
reshape (Reshape)                                                                      & (1,9)        & 0            \\
pixel\_softmax (PixelSoftmax)                                                          & (1,9)        & 0            \\ \hline
Total Trainable Parameters: 1,435,337                                                  &              &              \\ \hline
\end{tabular}
\vspace{0.2cm}
 \caption{\label{params}The layer-wise summary of the single U-Net used in our test and CEU-Net architecture for Pavia University with no patching. }
\end{table}

\paragraph{Experiment Configurations}

To test feature reduction and semantic segmentation techniques we do each combination of the techniques for each dataset. For the baseline AeroRIT U-Net we use the Pytorch library and their github\cite{c22}, for the rest we use the Keras library. 

For each dataset, we remove the background completely. We then do our feature reduction technique and reduce our feature size to 32 for 2D CAE and 30 for PCA and 3D CAE. These numbers were chosen as a tuned hyperparameter based on preliminary studies for each dataset.

The reduced data is then split into training and testing randomly five times for 5-fold cross validation with a testing percentage of 25\% for each semantic segmentation method.

For each dataset: 2D CAE trains for 100 epochs, and 3D CAE trains for 150 epochs, HybridSN, Single U-Net, and AeroRIT U-Net train for 150 epochs then ensemble U-Net trains for 200 epochs for each U-Net. 

Each semantic segmentation technique uses categorical cross entropy for the loss function and has a learning rate of 0.0001. To determine effectiveness of techniques, we use overall test accuracy.

\paragraph{Clustering Methods}
There were two clustering methods explored, K-Means++\cite{arthur2006k} and Gaussian Mixture Models (GMM)\cite{mclachlan1988mixture,maugis2009variable} clustering. K-Means uses the mean to calculate the centroid for each cluster, while GMM takes into account the variance of the data as well. Therefore, based on the distribution for each dataset, one method may work better than the other. In addition to clustering method, the number of clusters will also be varied in preliminary test to determine the most effective number of clusters for each dataset.

\section{Performance Comparison}
Here, our main goal is to compare the performance of CEU-Net on the original testing set to mini-batch SGD training and highlight how we can improve performance without using neighborhood information. We first briefly demonstrate that among various feature extraction approaches PCA provides the best performance. 

\begin{wraptable}{o}{0.5\textwidth}
\vspace{-0.5cm}
\caption{This table shows the results of our feature reduction techniques.}\label{feature}
\begin{tabular}{llll}
\hline
Dataset                               & PCA             & 2D-CAE & 3D-CAE \\ \hline
\multicolumn{1}{l|}{Indian Pines}     & \textbf{0.89}   & 0.5738 & 0.816  \\
\multicolumn{1}{l|}{Salinas}          & \textbf{0.9836} & 0.8987 & 0.9427 \\
\multicolumn{1}{l|}{Pavia University} & \textbf{0.9608} & 0.9204 & 0.9501 \\
\multicolumn{1}{l|}{KSC}              & \textbf{0.9501} & 0.9    & 0.9065 \\
\multicolumn{1}{l|}{Botswana}         & \textbf{0.9689} & 0.9068 & 0.9262
\end{tabular}
\vspace{0.2cm}
\end{wraptable}

\subsection{Feature Extraction}

The results in Table~\ref{feature} show that PCA is a superior feature reduction technique versus 2D and 3D CAE. PCA gets high overall accuracy in each pairing for all datasets. In addition, the feature reduction training time for 2D and 3D CAE are high as they are full neural networks, while PCA is very fast. Results of highest overall accuracy for each feature reduction technique can be seen in Table \ref{feature}. All reported results are overall test accuracy using our tuned CEU-Net for the classifier.

\subsection{Cluster Hyperparameters}
\begin{figure*}[t]
\vspace{4mm}
\centering
\includegraphics[scale=0.48]{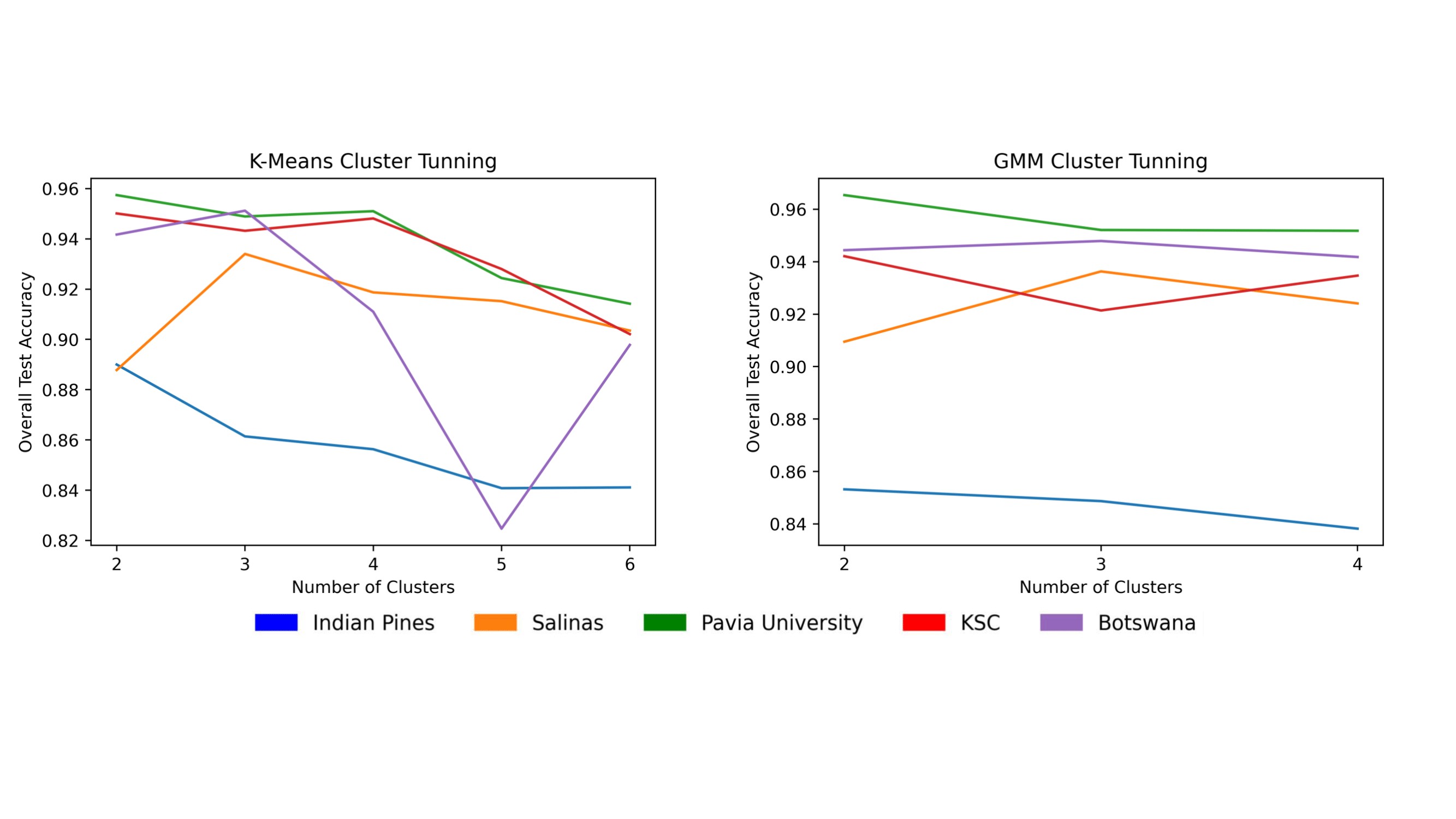}
\vspace*{-20mm}
\vspace{0.2cm}
\caption{This figure shows the results of our cluster tuning. We explored both K-Means and Gaussian Mixture Models (GMM) for our clustering methods along with a wide spread of cluster numbers. Any cluster larger than 4 for GMM or 6 for K-Means resulted in clusters with too little data for semantic segmentation in specific sub-U-Nets. The relatively small number of clusters in each dataset show how easily segmentatble these datasets are.}
\label{clustering}
\end{figure*}

The number of clusters and clustering method was considered a hyperparameter for each dataset for our CEU-Net. The results are shown in Figure \ref{clustering}. Therefore when the results show $X$\% overall accuracy in ensemble U-Net for Indian Pines, that was achieved through K-means with clusters as determined in Table \ref{no-patching} (k=2).

The preliminary test for these parameters involved testing out our CEU-Net with 5-fold cross validation using K-Means and GMM for cluster numbers 2-6. Due to the dataset sizes, often number of clusters $k > 4$ were impossible as too few samples would be sent to a single U-Net, this was dataset dependent. In addition, the performance of the U-Nets would drop significantly as shown in Figure \ref{clustering}.

\begin{figure}
\centering
\begin{minipage}{.48\columnwidth}
  \centering
  \includegraphics[scale=0.45]{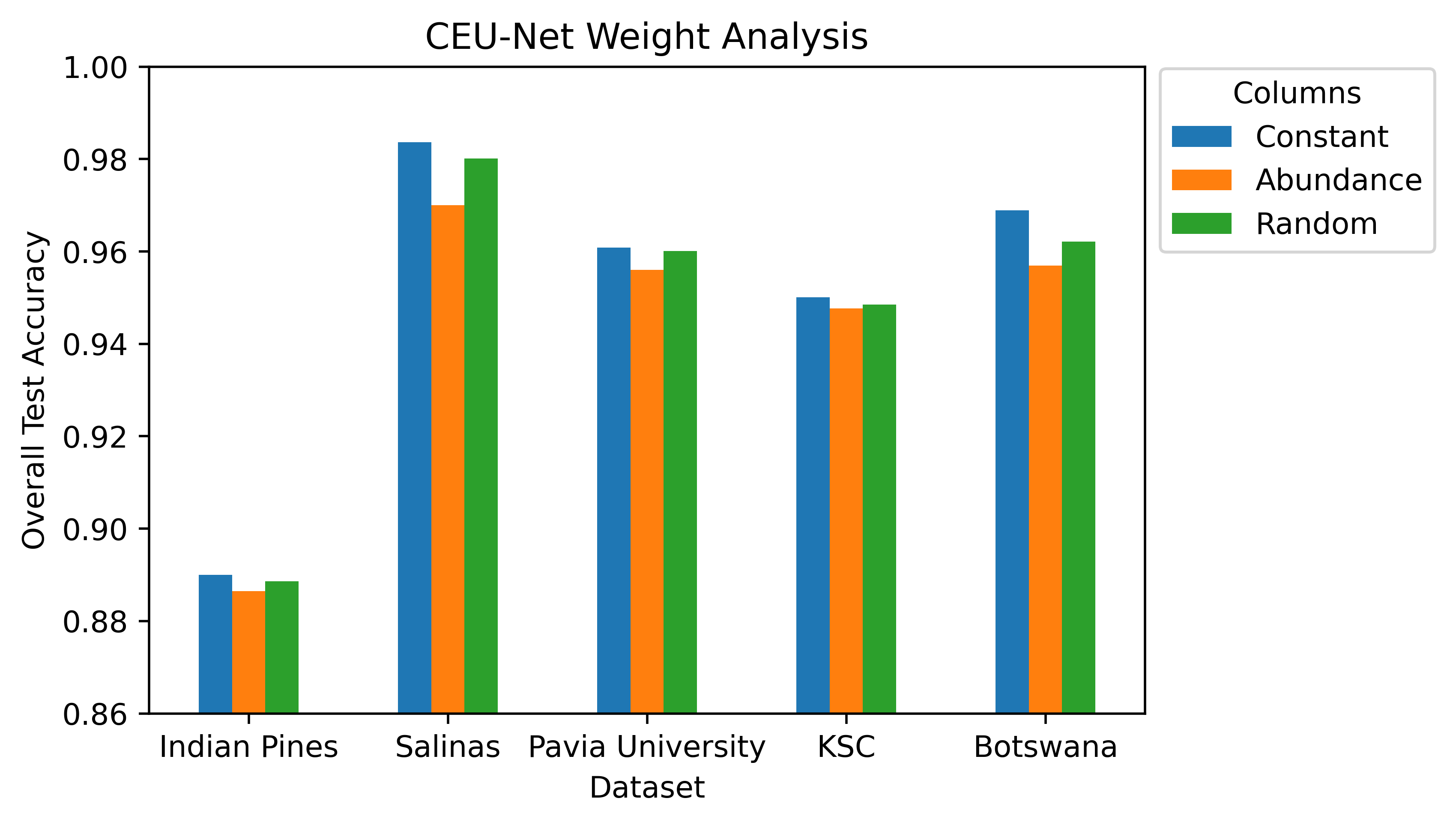}
  \caption{This figure shows the results of the loss weight $\omega=[\omega_1,\ldots,\omega_k]^T$ tuning for our ensemble CEU-Net method. Three different weight types are explored: 1) Constant weights in each sub-model, 2) Weights equaling the abundance of data given to each sub-model, 3) Random weights assigned. All weights have to sum to equal $1$. All test were run with 5-fold cross validation. We observer that the constant weights out perform other methods, therefore, we use constant weights in all of our CEU-Net experiments.}
  \label{weight}
\end{minipage}%
\hfill
\begin{minipage}{.5\columnwidth}
  \centering
  \includegraphics[scale=0.425]{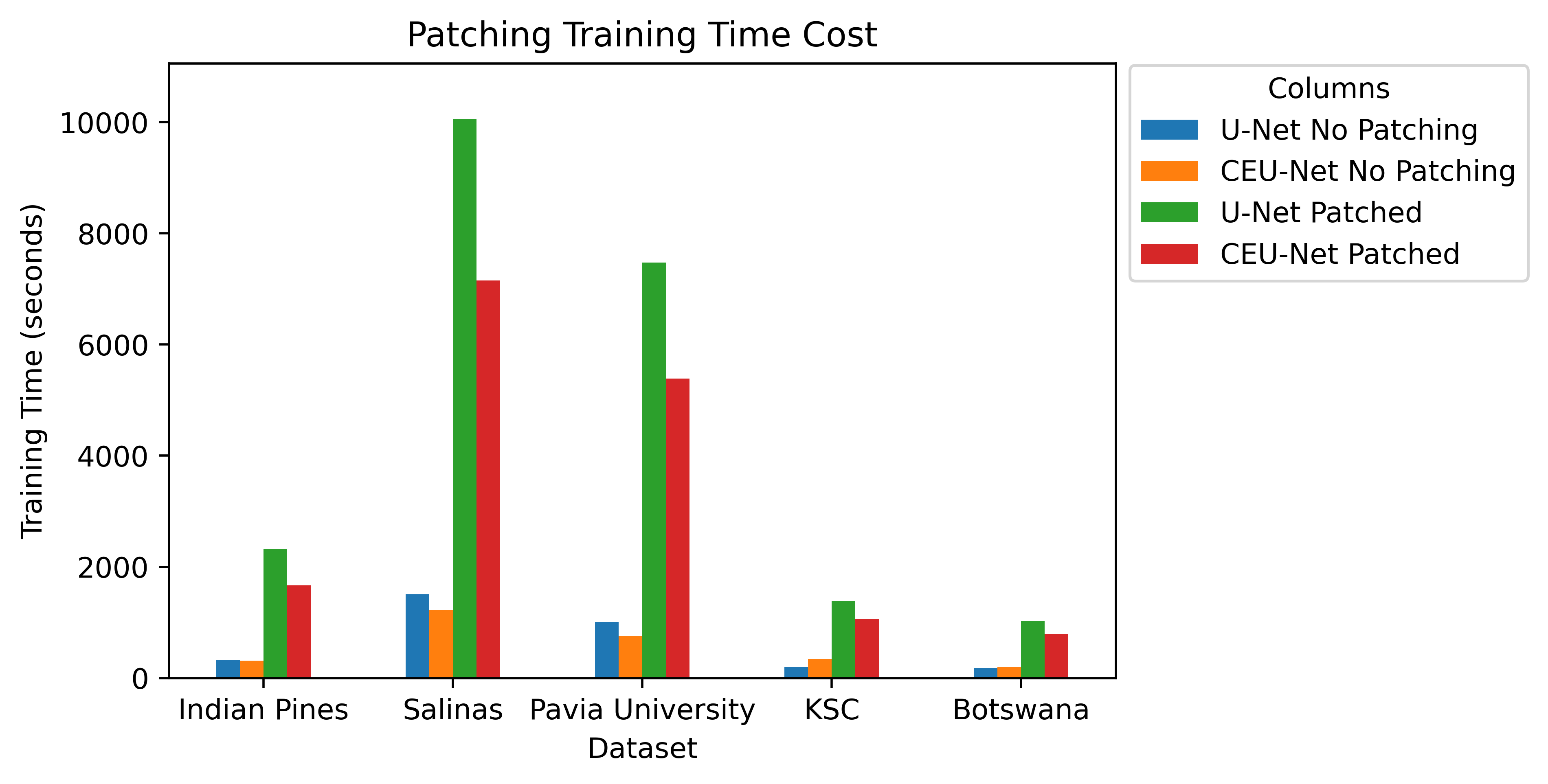}
 \caption{This figure shows the training time for U-Net and CEU-Net for each dataset. The optimal feature reduction technique was used before each semantic segmentation: Principal Component Analysis(PCA). Bands were reduced to 30. This figure shows the dramatic difference in training time when employing patching. Patching greatly increases the training time of semantic segmentation models. Center Pixel Classification (CPC) was used with a patch size of $10$ x $10$.}
  \label{time}
\end{minipage}
\end{figure}


\subsection{Weight Study}
Our CEU-Net is an ensemble method that uses a linear combination of the prediction of each sub single U-Net to give better overall accuracy on average than the single model. In ensemble networks, there are often sub-models that contributes more to an ensemble prediction and ones that are less skillful. Therefore, we tested different weighted loss average ensembles to analyze if giving attention/weight to certain sub-models is useful for CEU-Net in the context of HSI semantic segmentation. Equivalently, this experiment investigates the multipliers in the linear combinations of the each U-Net prediction presented in (\ref{eq:ensemble-loss2}).

For this study we execute three different ensemble manipulations:
\begin{enumerate}
    \item Constant weights: In this method we give equal attention to each sub-model in the CEU-Net by multiplying the loss of each by a constant.
    \item Abundance weights: In this method, the weight we use for the sub-model loss is equal to the percent pixel abundance that goes to each sub-model. Therefore, for example, if 60\% of the data goes to one model, the weight will be 0.6.
    \item Random weights: In this method the weights will be assigned at random as long as the sum of the weights equal to 1.
\end{enumerate}
Figure \ref{weight} shows that manipulating the attention of the ensemble network via loss weight modification has little bearing on over all test accuracy. We observe a constant loss weight modifier slightly increases overall test accuracy for CEU-Net in each dataset.

\subsection{Semantic Segmentation}

All models for each dataset trained for semantic segmentation without patching are shown in Table \ref{no-patching}. The optimal feature reduction technique was used before each semantic segmentation: Principal Component Analysis (PCA). Bands were reduced to 30. For each test, 5-fold cross validation was performed to ensure average result. We observe that our CEU-Net outperforms the baseline architectures and our single U-Net for each dataset.

 \begin{table}
 \centering
   \scalebox{1.1}{
\begin{tabular}{lllll}
\hline
Dataset                               & HybridSN & U-Net  & AeroRIT & CEU-Net        \\ \hline
\multicolumn{1}{l|}{Indian Pines}     & 0.8658   & 0.8701 & 0.7626        & \textbf{0.89 ($k=2$)}   \\
\multicolumn{1}{l|}{Salinas}          & 0.9679   & 0.9635 & 0.9625        & \textbf{0.9836 ($k=3$)} \\
\multicolumn{1}{l|}{Pavia University} & 0.9582   & 0.9592 & 0.9489        & \textbf{0.9608 ($k=2$)} \\
\multicolumn{1}{l|}{KSC}              & 0.9469   & 0.9485 & 0.9474        & \textbf{0.9501 ($k=2$)} \\
\multicolumn{1}{l|}{Botswana}         & 0.965    & 0.9677 & 0.9178        & \textbf{0.9689 ($k=3$)} \\ \hline
\end{tabular}
}
\caption{This table shows the overall test accuracy results for each semantic segmentation method for each dataset without patching.}
\label{no-patching}
\end{table}
\begin{table}
  \centering
  \scalebox{1.1}{
\begin{tabular}{llll}
\hline
Dataset                               & HybridSN        & U-Net           & CEU-Net        \\ \hline
\multicolumn{1}{l|}{Indian Pines}     & 0.9688          & \textbf{0.981}  & 0.980           \\
\multicolumn{1}{l|}{Salinas}          & 0.9983          & 0.9976          & \textbf{0.9987} \\
\multicolumn{1}{l|}{Pavia University} & 0.9948          & \textbf{0.9959} & 0.9915          \\
\multicolumn{1}{l|}{KSC}              & 0.9593          & 0.9676          & \textbf{0.9895} \\
\multicolumn{1}{l|}{Botswana}         & \textbf{0.9828} & 0.9741          & 0.9657          \\ \hline
\end{tabular}
}
\caption{This table shows the overall test accuracy results for each semantic segmentation method for each dataset while employing patching.}
\label{patching}
\end{table}

As our second semantic segmentation study, we investigate how CEU-Net performs with patching relative to the baseline. Results are shown in Table \ref{patching}. The optimal feature reduction technique was used before each semantic segmentation: Principal Component Analysis(PCA). Bands were reduced to 30. Center Pixel Classification (CPC) was used with a patch size of $10$ x $10$. Our network appears to out perform the baseline in most datasets used.

Seeing the results of Tables \ref{no-patching} and \ref{patching}, patching appears to increase overall test accuracy each time. However, the cost of patching is dramatic in training time. Figure \ref{time} shows the exponential increase in training time that a relatively small patch size of $10$ x $10$ creates.

\section{Discussions}

\subsection{Patching}

For Indian Pines, we lose quite a bit of accuracy when compared to the 98\% accuracy that our CEU-Net with patching reports. However, our CEU-Net out performed all other models with an accuracy of 89\%. All other datasets show overall accuracies closer to their patched counterparts with accuracies above 90\%. Once the clustering method is an unsupervised non-deterministic method that does not use neighborhood information, we expected smaller accuracy versus patching for these datasets, however, our method can be used in datasets where patching is not as useful. As we observe, patching dramatically increases runtime even with a smaller patch size of $10$ x $10$. In comparison to our baseline models, HybridSN uses a $25$ x $25$ patch size\cite{c12} and AeroRIT uses a $64$ x $64$ patch size\cite{c22}, which increases training time significantly.

\subsection{Feature Reduction}

Out of all the pairs, it appears that Principal Component Analysis (PCA) is the better feature reducer when compared to neural network autoencoder approaches. Most papers that used autoencoders used neighborhood information in the feature reduction process via patching, making the information used not purely spectral information\cite{c23}. Our experimental results show that PCA outperforms autoencoders in overall test accuracy when only spectral information is considered. 

\subsection{Semantic Segmentation}

U-Net has seen great results for semantic segmentation on medical imagery and it appears to work well in the hyperspectral domain as well. Single U-Net out performed each baseline in all datasets when not patching them.

In addition to our single U-Net's success, we extend it with CEU-Net which out performs single U-Net in every dataset. This proves that it is not only possible to cluster pixels by their spectral signatures without knowing their individual class, but that it out performs single network models as well. In addition, CEU-Net has better training time compared to single U-Net in all datasets with patching and without patching, with the exception of Botswana and KSC without patching. U-Net performs better in training time for Botswana and KSC without patching due to the limited number of labeled samples allowing the overhead of training the clustering method in CEU-Net to dominate its training time. This issue would be remedied with larger datasets like Salinas. 

CEU-Net model performs better than Single U-Net and baselines in overall test accuracy in all datasets with no patching, as shown in Table \ref{no-patching}. With Patching, CEU-Net and Single U-Net both out-preform the baseline in four out of five datasets, as shown in Table \ref{patching}. Therefore, CEU-Net is a superior semantic segmentation model due to its superior spectral domain classification.

Our proposed CEU-Net increases overall accuracy by splitting the dataset into similar pixels via unsupervised clustering. This way, each sub-model can become an expert in these like pixels, allowing the network to detect minuscule differences between them that more generalized networks might miss, thereby increasing test accuracy. Then, combining the results of each specialized sub-model results in an overall accuracy larger than an individual model can achieve. The strength of CEU-Net is further increased with the addition of patching if the dataset benefits. 

\section{Conclusion}

Before this paper, there has not been a proper investigation, to our knowledge, on the role neighborhood information should have in HSI semantic segmentation. Through our discussion, we showed the weaknesses of patching for complex datasets, but also its strengths under particular conditions. By exploring feature reduction and semantic segmentation techniques without using neighborhood information, our single U-Net and CEU-Net achieves competitive accuracies without it. We also debuted a novel network called CEU-Net that out performs all baselines with a preprocessing step that is dataset independent, unsupervised, and does not use neighborhood information. We believe that Clustering Ensemble U-Net can be used in future works on many datasets, especially ones that are extra challenging with complex and overlapping class labels where neighborhood information is weak. In addition, we showed CEU-Net also outperforms the baseline networks like HybridSN with patching as a preprocessing step, showing that CEU-Net is an excellent general purpose HSI semantic segmentation technique.

\section*{Acknowledgment}
This work has been partially supported  NSF 1920908 (EPSCoR RII Track-2); the findings are those of the authors only and do not represent any position of these funding bodies.

\bibliographystyle{unsrt}  

\bibliography{main}

\end{document}